%% file: main.tex
\documentclass[10pt,twocolumn,letterpaper]{article}

\usepackage{wacv}              %

\usepackage[accsupp]{axessibility}

\usepackage{graphicx}
\usepackage{times}
\usepackage{amsmath}
\usepackage{amssymb}
\usepackage{booktabs}

\newcommand{\spm}[1]{{\tiny$\pm$#1}}
\DeclareMathOperator*{\argmax}{argmax}
\DeclareMathOperator*{\softmax}{softmax}

\usepackage{pgfplots}
\pgfplotsset{compat=1.18}
\usepgfplotslibrary{groupplots}
\usepackage{booktabs}
\usepackage{tikz}
\usetikzlibrary{shapes.geometric, positioning, calc, fit}

\usepackage{algorithm}
\usepackage{algpseudocode}
\usepackage{dsfont}

\usepackage{multirow}

\usepackage{arydshln}

\makeatletter
\def\adl@drawiv#1#2#3{%
        \hskip.5\tabcolsep
        \xleaders#3{#2.5\@tempdimb #1{1}#2.5\@tempdimb}%
                #2\z@ plus1fil minus1fil\relax
        \hskip.5\tabcolsep}
\newcommand{\cdashlinelr}[1]{%
  \noalign{\vskip\aboverulesep
           \global\let\@dashdrawstore\adl@draw
           \global\let\adl@draw\adl@drawiv}
  \cdashline{#1}
  \noalign{\global\let\adl@draw\@dashdrawstore
           \vskip\belowrulesep}}
\makeatother

\usepackage[pagebackref,breaklinks,colorlinks]{hyperref}

\usepackage[capitalize]{cleveref}
\crefname{section}{Sec.}{Secs.}
\Crefname{section}{Section}{Sections}
\Crefname{table}{Table}{Tables}
\crefname{table}{Tab.}{Tabs.}

\def\wacvPaperID{1488} %
\def\confName{WACV}
\def\confYear{2024}

\begin{document}

\title{Improving Open-Set Semi-Supervised Learning with Self-Supervision}

\author{Erik Wallin\textsuperscript{1,2}, Lennart Svensson\textsuperscript{2}, Fredrik Kahl\textsuperscript{2}, Lars Hammarstrand\textsuperscript{2} \\
\textsuperscript{1}Saab AB, \textsuperscript{2}Chalmers University of Technology \\
\{walline,lennart.svensson,fredrik.kahl,lars.hammarstrand\}@chalmers.se}

\maketitle

\begin{abstract}
Open-set semi-supervised learning (OSSL) embodies a practical scenario within semi-supervised learning, wherein the unlabeled training set encompasses classes absent from the labeled set.
Many existing OSSL methods assume that these out-of-distribution data are harmful and put effort into excluding data belonging to unknown classes from the training objective. In contrast, we propose an OSSL framework that facilitates learning from all unlabeled data through self-supervision. %
    Additionally, we utilize an energy-based score to accurately recognize data belonging to the known classes,
    making our method well-suited for handling uncurated data in deployment.
    We show through extensive experimental evaluations that our method yields state-of-the-art results on many of the evaluated benchmark problems in terms of closed-set accuracy and open-set recognition when compared with existing methods for OSSL.
    Our code is available at \url{https://github.com/walline/ssl-tf2-sefoss}.
\end{abstract}

\section{Introduction}

Using a combination of labeled and unlabeled data for training a model, so-called \emph{semi-supervised learning} (SSL), is a well-studied field of machine learning \cite{scudder1965probability, yarowsky1995unsupervised, lee2013pseudo, tarvainen2017mean, sohn2020fixmatch}, motivated by exploiting the extensive amounts of cheap and readily available unlabeled data. %
However, semi-supervised learning is typically studied in a \emph{closed-set context}, where labeled and unlabeled data are assumed to follow the same distribution. In practice though, one can expect that the labeled set is of a much more curated character, \eg, hand-picked examples from known classes, compared to its unlabeled counterpart, which may contain outliers or corrupted data.
Semi-supervised learning where the unlabeled set contains more classes than the labeled set (see \cref{fig:introfig}) is referred to as \emph{open-set semi-supervised learning} (OSSL), where classes present in the labeled set are considered in-distribution (ID), whereas other classes are recognized as out-of-distribution (OOD).

A common approach for training models in OSSL is to use a standard SSL objective but only include unlabeled data that are predicted as ID \cite{chen2020semi, guo2020safe, he2022safe, yu2020multi}, while the rest are, \eg, discarded \cite{chen2020learning} or given less importance \cite{guo2020safe}. This approach is motivated by an assumption that the training signal from OOD data is harmful. %
Consequently, by restricting learning to samples predicted as ID they are not efficiently using all available data. 
Some algorithms take steps towards better employment of unlabeled data in OSSL. For example, \cite{saito2021openmatch} uses consistency regularization and \cite{huang2021trash} introduces a self-supervised rotation loss on all unlabeled data, while \cite{huang2022they} identifies OOD data semantically similar to ID data that can be ``recycled'' as such.

\begin{figure}[!t]
    \centering
    \input{figures/introfig-ossl}
    \vspace*{-0.2cm}
    \caption{Comparison between open- and closed-set semi-supervised learning. %
    }
    \label{fig:introfig}
    \vspace*{-0.55cm}
\end{figure}

In this work, we propose a \textbf{Se}lf-supervision-centric \textbf{F}ramework for \textbf{O}pen-set \textbf{S}emi-\textbf{S}upervised learning (SeFOSS). Our framework unconditionally promotes learning from all unlabeled data by utilizing a self-supervised consistency loss, %
which is effective in the traditional closed-set SSL setting \cite{wallin2022doublematch}. Compared to previous methods focusing on predicted inliers through an SSL objective, we argue that making self-supervision the primary source of learning is key, both for data efficiency and robustness in OSSL. The impact of classifying unlabeled data incorrectly as ID, or as the incorrect class, is intuitively much less significant for a self-supervised training objective. To our knowledge, we are the first to utilize self-supervised feature consistency of the type used in \cite{wallin2022doublematch} for OSSL.
Moreover, we prioritize the final model's capacity for open-set recognition (OSR) (predicting data as ID or OOD). To this end, we resort to the recently popularized \emph{free-energy score} \cite{li2020energy}. This theoretically founded approach for OSR is well-performing with the added benefit of not requiring any architectural modifications to the classification model.

We assess our framework and compare it with existing approaches through extensive experiments across diverse datasets.
The results show that SeFOSS reaches state-of-the-art performance for closed-set accuracy and OSR on many of the evaluated benchmark problems. Although SeFOSS does not reach SOTA on all problems, it consistently displays strong results across the range of evaluated scenarios. We also show that the training of SeFOSS is stable, avoiding the need for additional validation data for early stopping. Furthermore, our experiments reveal that the previous assumption \cite{guo2020safe, chen2020semi, he2022safe} that OOD data significantly hurt the closed-set accuracy of traditional SSL is not always valid. On the contrary, the seminal SSL method FixMatch \cite{sohn2020fixmatch} outperforms all evaluated OSSL methods on closed-set accuracy. Thus, we suggest that designated OSSL methods are mainly important when OSR performance is vital.

Our main contributions are:
\begin{itemize} \vspace*{-0.15cm}
    \item a framework for OSSL that incorporates self-supervised feature consistency for accurate closed-set performance and OSR \vspace*{-0.15cm}
    \item an extensive evaluation across a wide range of open-set scenarios showing that our framework achieves strong results compared to existing methods. \vspace*{-0.15cm}
    \item a challenge to the prior assumption that OOD data in the unlabeled set significantly harms the closed-set performance of traditional SSL methods, indicating that research efforts should be put elsewhere. \vspace*{-0.15cm}
\end{itemize}

\section{Related work}

\textbf{Semi-supervised learning:} Research in SSL has a long history \cite{scudder1965probability, fralick1967learning,agrawala1970learning, yarowsky1995unsupervised,lee2013pseudo,rasmus2015semi}. Recently, frameworks such as FixMatch \cite{sohn2020fixmatch} and UDA \cite{xie2019unsupervised} introduced a new paradigm for SSL combining pseudo-labeling with consistency regularization using data augmentations. These works emphasize the importance of strong domain-specific data augmentations for high performance in SSL. In the image domain, these augmentations are, \eg, RandAugment \cite{cubuk2020randaugment} and Cutout \cite{devries2017improved}. The effectiveness of FixMatch and UDA sparked a new wave of research trying to extend or improve these frameworks. For example, FlexMatch \cite{zhang2021flexmatch}, Dash \cite{xu2021dash}, SimMatch \cite{zheng2022simmatch}, CCSSL \cite{yang2022class}, DP-SSL \cite{xu2021dp}, and DoubleMatch~\cite{wallin2022doublematch} all propose ways to improve the strategies for pseudo-labeling and consistency regularization of UDA and FixMatch.

For this work, we take particular inspiration from DoubleMatch~\cite{wallin2022doublematch}, which highlighted the effectiveness of enabling learning from \emph{all} unlabeled data. DoubleMatch is motivated by the fact that UDA and FixMatch restrict learning from unlabeled data to samples for which the model produces confident predictions. In this regard, DoubleMatch adds a self-supervised cosine-similarity loss on the feature predictions across augmentations of all unlabeled data. By
promoting prediction consistency also for uncertain unlabeled data,
DoubleMatch sees improvements in terms of both final accuracy and training speed. In the context of OSSL, we suggest that this ability to safely learn from all unlabeled data, without inferring class predictions, is of particular interest as it can enable the model to learn from outlier data. For this reason, we include the self-supervision proposed by DoubleMatch as a core part of SeFOSS.

\textbf{Open-set recognition:} The ability to identify previously unseen classes is an important safety feature in many machine learning applications and the task of predicting if data belongs to a pre-defined set of classes or not is often referred to as open-set recognition. This problem is widely studied \cite{scheirer2012toward, bendale2016towards, hendrycks2016baseline, hendrycks2018deep, lakshminarayanan2017simple, lee2018simple, liang2017enhancing} where existing solutions are based on, \eg, modeling class conditional feature distributions \cite{lee2018simple}, using ensembles of models \cite{lakshminarayanan2017simple}, or analyzing the model predictions under perturbations of input data \cite{liang2017enhancing}.

Recently, Li \etal \cite{li2020energy} proposed to use the \emph{free-energy score} for OSR, which for a data point $\mathbf{x}$ is obtained by interpreting the logits $f_y(\mathbf{x})$ for each class $y$ as negative energies: $E(\mathbf{x},y) = -f_y(\mathbf{x})$. The free-energy score is then given by
\begin{equation}
    \label{eq:free-energy}
    F(\mathbf{x}) = -\frac{1}{\beta}\log \sum_{y^\prime=1}^C e^{-\beta E(\mathbf{x},y^\prime)},
\end{equation}
where $\beta$ is a hyperparameter and $C$ is the number of classes. The free-energy score is theoretically aligned with the marginal distribution for ID data, $p(\mathbf{x})$, so that we can expect $F(\mathbf{x}_\text{a}) < F(\mathbf{x}_\text{b})$ for an $\mathbf{x}_\text{a}$ that is ID and an $\mathbf{x}_\text{b}$ that is OOD. It's worth noting that, for large $\beta$, we get $F(\mathbf{x}) \approx \min_{y^\prime} E(\mathbf{x}, y^\prime)$, \ie., the maximum logit score, which also has been used successfully for OSR \cite{vaze2021open}.

For SeFOSS, we utilize the free-energy score to determine which samples are ID or OOD. Our main motivation is its simplicity, \ie, not requiring architectural modifications or significant computational complexity, while still being a powerful discriminant.
Another benefit of using a method that can be ``plugged in'' to any existing model is that it allows for easy and fair comparisons of results.

\textbf{Open-set semi-supervised learning:} In OSSL \cite{chen2020semi, yu2020multi, he2022safe, huang2021trash, saito2021openmatch, huang2022they, wang2023out, guo2020safe, huang2023fix, he2022not, park2022opencos, mo2023ropaws, han2023pseudo, zhao2022out}, we study the case where the unlabeled training set, and sometimes also the test set, contains additional classes not present in the labeled set. 
Compared to (closed-set) SSL, this is a more general and much less studied problem, and compared to \emph{open-set domain adaptation} (OSDA) \cite{panareda2017open, saito2018open}, there are no assumptions regarding domain shifts in the training sets. 

Many proposed methods for OSSL classify the unlabeled data as either ID or OOD and use only confident ID samples for the training objective in a traditional SSL scheme. One such example is {UASD} \cite{chen2020semi}, where unlabeled data are classified as inliers based on thresholding the maximum of predicted softmax distribution.
The method averages multiple predictions of the same unlabeled data from different time steps during training for increased predictive calibration.
The framework {MTCF} \cite{yu2020multi} uses a similar strategy but employs a separate OSR head for predicting the probability of a sample being ID. The OSR head trains in an SSL fashion where the OSR head uses model predictions as optimization targets.
In {SAFE-STUDENT} \cite{he2022safe}, the model is trained using both pseudo-in- and outliers, which are predicted from unlabeled data by using %
energy discrepancy.

{OpenMatch} \cite{saito2021openmatch} and {T2T} \cite{huang2021trash} use a slightly different strategy built around a one-vs-all framework. In both cases, we have one head for each class predicting the probability of a sample belonging to the corresponding class. Predicted inliers are used for SSL objectives based on FixMatch (OpenMatch) or UDA (T2T).
Both T2T and OpenMatch take steps towards utilizing all unlabeled data: OpenMatch through a consistency loss for its one-vs-all predictions, and T2T through a self-supervised rotation loss.

In the {DS$^\mathbf{3}$L} method \cite{guo2020safe}, the focus is on preserving closed-set performance by solving the OSSL problem via a bi-level optimization. The inner task is to learn the classifier based on a standard two-term SSL optimization but scaling the loss for unlabeled samples using a data-dependent weight function. The outer task is thus to learn the weight function that minimizes the \emph{labeled} loss term of the inner task. This way, the outer optimization steers the training by using the labeled set as a proxy so that the model never drops in closed-set performance.

Lastly, {TOOR} \cite{huang2022they} proposes to identify ``recyclable'' OOD data, \ie, semantically close to one of the ID classes.
The method projects the recyclable OOD data on the space of ID data by domain adaptation and uses them in conjunction with unlabeled data predicted as ID for training through an SSL objective.

In summary, most existing contributions for OSSL are based on the assumption that OOD data are ``harmful'' and focus on detecting ID data from the unlabeled set to use in an SSL training objective. In SeFOSS, we instead facilitate learning from all unlabeled data while also learning to distinguish between ID and OOD. Additionally, in contrast to many prior works, we do not require extra model heads for classifying data as ID or OOD. Using the free energy score, we avoid additional model parameters or heavy computational complexity for solving the OSR task.

\section{Method}

This section describes SeFOSS, our proposed method for OSSL. The main philosophy behind this method is that we encourage learning from \emph{all} unlabeled data, whether it is ID or OOD. Our proposed method achieves this by applying the self-supervised loss of DoubleMatch~\cite{wallin2022doublematch} on all unlabeled samples in each training batch.

We complement the self-supervision with losses on unlabeled data confidently predicted as ID or OOD to improve OSR performance. For samples confidently predicted as ID, we apply a pseudo-labeling loss similar to those of FixMatch \cite{sohn2020fixmatch} and UDA \cite{xie2019unsupervised}. For unlabeled data confidently predicted as outliers, we instead use energy regularization to increase the model's confidence that these are OOD. \Cref{fig:method} illustrates how SeFOSS treats unlabeled data.

Following standard SSL practice, losses on unlabeled data are combined with supervised cross-entropy on labeled data. The components of SeFOSS are detailed below.

\begin{figure*}
    \centering
    \input{figures/method-graph}
    \vspace*{-0.2cm}
    \caption{Sorting of \emph{unlabeled data} in SeFOSS. Predicted outliers are used for energy regularization in $l_e$. Predicted inliers are used for a pseudo-labeling loss in $l_p$. All unlabeled data are used for self-supervision in $l_s$.}
    \label{fig:method}
    \vspace*{-0.5cm}
\end{figure*}

\subsection{Self-supervision on all unlabeled data}

The central source of learning from unlabeled data in SeFOSS is self-supervision.
To this end, we use the loss proposed by \cite{wallin2022doublematch}, \ie, a cosine similarity between feature predictions for different augmentations of unlabeled data:
\begin{equation}
    \label{eq:ls}
    l_s = - \frac{1}{\mu B} \sum_{i=1}^{\mu B} \frac{h(\mathbf{v}_i) \cdot \mathbf{z}_i} {\|h(\mathbf{v}_i)\| \|\mathbf{z_i}\|},
\end{equation}
where $\mu B$ is the unlabeled batch size, $\mathbf{v}_i$ and $\mathbf{z}_i$ are $d$-sized feature vectors from the penultimate network layer for weak and strong augmentations of sample $i$, respectively. The operator $\| \cdot \|$ is the $l_2$ norm. The mapping $h : \mathds{R}^d \rightarrow \mathds{R}^d$ is a trainable linear projection to allow for differences in feature predictions for weak and strong augmentations. In gradient evaluations of $l_s$, we treat $\mathbf{z}_i$ as constant.

A principal difference between the self-supervision of \eqref{eq:ls} and the losses for all unlabeled data in T2T \cite{huang2021trash} and OpenMatch \cite{saito2021openmatch}, is that \eqref{eq:ls} makes use of strong data-augmentation whereas the corresponding losses of T2T and OpenMatch use weak data-augmentation only.

\subsection{Pseudo-labeling loss for pseudo-inliers}

SeFOSS uses the free-energy score \cite{li2020energy} as defined in \eqref{eq:free-energy} for OSR.
For convenience, we define the equivalent function $s : \mathds{R}^C \rightarrow \mathds{R}$ that operates on the logits $\boldsymbol{\sigma}$ as
\begin{equation}
    \label{eq:s-score}
    s(\boldsymbol{\sigma}) = - \log \sum_{y^\prime = 1}^C e^{\sigma_{y\prime}},
\end{equation}
where $\sigma_{y^\prime} = f_{y^\prime}(\mathbf{x})$ is the logit associated with class $y^\prime$ for data point $\mathbf{x}$. For data that are confidently ID (pseudo-inliers), we expect $s(\boldsymbol{\sigma})$ to be low. To amplify this confidence, we apply the pseudo-labeling loss
\begin{equation}
    \label{eq:lp}
    \begin{split}
        l_p = \frac{1}{\mu B} \sum_{i=1}^{\mu B} &\mathds{1} \{ s(\mathbf{w}_i) < \tau_\text{id} \} \\ &\times H(\argmax(\mathbf{w}_i),  \softmax(\mathbf{q}_i)),
    \end{split}
\end{equation}
where $\tau_\text{id}$ is a confidence threshold. We let $\argmax~:~\mathds{R}^C~\rightarrow~\mathds{R}^C$ so that it returns a one-hot vector. Similarly, we define $\softmax : \mathds{R}^C \rightarrow \mathds{R}^C$ so that the $j$:th element is
\begin{equation}
    \label{eq:softmax}
    \softmax(\boldsymbol{\sigma})_j = \frac{e^{\sigma_j}}{\sum_{y^\prime=1}^C e^{\sigma_{y\prime}}}, \; \text{for}  \; j=1, \dots, C.
\end{equation}
The inputs $\mathbf{w}_i$ and $\mathbf{q}_i$ in \eqref{eq:lp} are the predicted logits for unlabeled sample $i$ for weak and strong data augmentations, respectively. Lastly, $\mathds{1}\{\cdot\}$ is the indicator function and $H(\cdot, \cdot)$ is the cross entropy between two discrete probability distributions $\mathbf{p}^a$, $\mathbf{p}^b$ given by
\begin{equation}
H(\mathbf{p}^a,\mathbf{p}^b) = - \sum_{i=1}^C p_i^a \log p_i^b,
\end{equation}
where $p_i^a$ and $p_i^b$ are the $i$:th elements of $\mathbf{p}^a \in \mathds{R}^C$ and $\mathbf{p}^b \in \mathds{R}^C$, respectively. As for $l_s$ from \eqref{eq:ls}, we consider the predictions on weakly augmented data as constant when evaluating the gradient of $l_p$.

Note that $l_p$ is equivalent to the pseudo-labeling loss in FixMatch \cite{sohn2020fixmatch} with the exception that FixMatch selects pseudo-labels by thresholding the maximum value of the predicted softmax distributions. We choose pseudo-labels by thresholding the free-energy score, which is a better score for OSR than the maximum softmax probability \cite{li2020energy}.

\subsection{Energy regularization for pseudo-outliers}

Many existing methods for OSSL \cite{chen2020semi,guo2020safe,yu2020multi} focus on identifying unlabeled data that are confidently ID. To boost the separation of OSR predictions between ID and OOD, we suggest including a loss for unlabeled data confidently predicted as OOD, \ie, pseudo-outliers. %
Inspired by \cite{li2020energy}, we employ a hinge loss on the free-energy score of the pseudo-outliers to stimulate the model to raise the free-energy score to a given margin. This energy regularization is given by
\begin{equation}
    \label{eq:le}
    l_e = \frac{\sum_{i=1}^{\mu B} \mathds{1} \{ s(\mathbf{w}_i) > \tau_\text{ood} \} \max(0, m_\text{ood} - s(\mathbf{w}_i))^2}{\ \sum_{i=1}^{\mu B} \mathds{1} \{ s(\mathbf{w}_i) > \tau_\text{ood} \}},
\end{equation}
where $\tau_\text{ood} \in \mathds{R}$ is the confidence threshold for OOD data and $m_\text{ood} \in \mathds{R}$ is the margin for the hinge loss. Note that $l_e$ only uses the predictions for weakly augmented data because we empirically found that using strongly augmented data in $l_e$ led to instabilities during training.

\subsection{Adaptive confidence thresholds}

In SeFOSS, we select pseudo-in- and outliers from unlabeled data, cf.\ \eqref{eq:lp} and \eqref{eq:le}, based on thresholding the free-energy score $s$ defined in \eqref{eq:s-score}. The free-energy score is non-probabilistic and thus unbounded, so selecting these thresholds is not trivial.
Thus, we propose adaptively calculating $\tau_\text{id}, \tau_\text{ood}$ and $m_\text{ood}$ based on the distribution of $s$ given our labeled training set at the end of a pre-training phase.
During this pre-training phase, the model is trained only with a supervised loss on labeled data and the self-supervised loss given by \eqref{eq:ls} on unlabeled data. 
Following the pre-training phase, we compute $s$ for the complete (unaugmented) labeled training set.
The energy scores are evaluated using an exponential moving average of the model parameters for stability.
Given the set of energy scores $\{ S_i : i = 1, \dots, M \}$, where $M$ is the number of labeled training data, we compute the median $S_m$ and the interquartile range $S_{iqr}$. The confidence thresholds and the margin for the energy regularization are then set as
\begin{equation}
    \label{eq:thresholds}
    \begin{split}
        \tau_\text{id} &\leftarrow S_m - S_{iqr} \cdot \zeta_\text{id}, \\
        \tau_\text{ood} &\leftarrow S_m + S_{iqr} \cdot \zeta_\text{ood} \\
        m_\text{ood} &\leftarrow S_m + S_{iqr} \cdot \xi_\text{ood},
    \end{split}
\end{equation}
where $\zeta_\text{id}$, $\zeta_\text{ood}$, and $\xi_\text{ood}$ are scalar hyperparameters. By using these adaptive metrics, we expect a tuned set of the hyperparameters $\zeta_\text{id}$, $\zeta_\text{ood}$, and $\xi_\text{ood}$ to work for a wider set of problems, compared to direct tuning of $\tau_\text{id}$, $\tau_\text{ood}$, and $m_\text{ood}$. Compared to the Otsu method \cite{otsu1979threshold} used for adaptive thresholding on unlabeled data in OSSL \cite{huang2021trash, wang2023out}, our method has no requirement for clearly bi-modal situations \cite{kittler1985threshold} since we only consider ID training data. Moreover, in contrast to the adaptive thresholds for pseudo-labeling used in closed-set SSL \cite{zhang2021flexmatch, xu2021dash, wang2022freematch}, their purpose is to lower thresholds to assign more pseudo-labels, which is not desired in OSSL.

\subsection{Full training objective}

The full training loss is a weighted sum of five terms:
\begin{equation}
    \label{eq:final-loss}
    l = l_l + w_p l_p + w_s l_s + w_e l_e + w_w l_w, 
\end{equation}
where $w_p$, $w_s$, $w_e$, and $w_w$ are scalar hyperparameters for controlling the relative importance of each of the terms. Here, $l_s$ is the self-supervision loss in \eqref{eq:ls}, $l_p$ is the pseudo-labeling loss in \eqref{eq:lp}, and $l_e$ is the energy regularization term in \eqref{eq:le}. We also use a supervised loss for the labeled data:
\begin{equation}
    \label{eq:ll}
    l_l = \frac{1}{B} \sum_{i=1}^{B} H(\mathbf{y}_i, \softmax(\mathbf{o}_i)),
\end{equation}
where $B$ is the number of labeled data in each batch, $\mathbf{y}_i \in \mathds{R}^C$ is the one-hot label vector for labeled sample $i$, and $\mathbf{o}_i \in \mathds{R}^C$ is the predicted logits for weakly augmented labeled sample $i$. Lastly, we add weight-regularization
\begin{equation}
    \label{eq:weight-decay}
    l_w = \frac{1}{2} \| \boldsymbol{\theta} \|^2, 
\end{equation}
$\boldsymbol{\theta}$ being a vector of all trainable weights (excluding biases). %

\begin{figure}[t!]
\begin{algorithm}[H]
    \centering
    \footnotesize
    \input{algorithms/trainingloop}
    \caption{SeFOSS training loop}
    \label{alg:trainingloop}
\end{algorithm}
\vspace{-.8cm}
\end{figure}
\begin{figure}[h!]
\begin{algorithm}[H]
    \centering
    \footnotesize
    \input{algorithms/energymatch}
    \caption{SeFOSS training step}
    \label{alg:energymatch}
\end{algorithm}
\vspace{-.8cm}
\end{figure}

\subsection{Data augmentation and optimization}

\begin{figure*}[t]
    \centering
    \input{figures/data-examples}%
    \vspace*{-0.15cm}
    \caption{A few representative examples from the datasets used in our experiments.}
    \label{fig:data-examples}
    \vspace*{-0.35cm}
\end{figure*}

Our method utilizes weak or strong data augmentation for labeled and unlabeled data during training, where we follow one of the proposed augmentation strategies from FixMatch \cite{sohn2020fixmatch}. Weak augmentations involve a stochastic flip-and-shift strategy. Strong augmentation stacks the weak augmentation with two randomly selected transformations from RandAugment \cite{cubuk2020randaugment}, followed by Cutout \cite{devries2017improved}.

For optimization, we use stochastic gradient descent with Nesterov momentum \cite{sutskever2013importance}. We define a scheme for the learning rate $\eta$ where the learning rate stays constant in the pre-training phase and follows a cosine decay in the subsequent training phase:
\begin{equation}
    \label{eq:lr}
    \eta(k) = 
    \begin{cases}
    \eta_0 \quad &\text{for} \quad k < K_p \\
    \eta_0 \cos \left( \gamma \frac{\pi (k-K_p)}{2 (K-K_p)} \right) \quad &\text{otherwise} %
    \end{cases},
\end{equation}
where $\eta_0$ is the initial learning rate, $\gamma$ is a hyperparameter that controls the decay rate, $k$ is the current training step, $K_p$ is the number of pre-training steps, and $K$ is the total number of training steps.

Our training procedure is summarized in \cref{alg:trainingloop}, and each training step is detailed in \cref{alg:energymatch}, where we denote the trainable parts of our model as $f$, $g$, and $h$. The backbone model $f : \mathds{R}^D \rightarrow \mathds{R}^d$ maps the input images of dimension $D$ to the feature predictions of dimension $d$. The classification head, $g : \mathds{R}^d \rightarrow \mathds{R}^C$ is a linear head that predicts logits given feature predictions. Finally, $h : \mathds{R}^d \rightarrow \mathds{R}^d$ is a projection head that performs a linear transformation on the feature predictions of strongly augmented data to the feature space of weakly augmented data, cf.\ \eqref{eq:ls}.

\section{Experiments}

We present experimental results for SeFOSS alongside other SSL and OSSL methods across various datasets. The OSSL methods that we compare with, MTCF \cite{yu2020multi}, OpenMatch \cite{saito2021openmatch}, and T2T \cite{huang2021trash}, all seek to produce well-performing models in terms of both closed-set accuracy and OSR. For comparison, we include closed-set SSL method FixMatch \cite{sohn2020fixmatch} and a fully supervised baseline trained using only the labeled subset. %
We evaluate closed-set accuracy and OSR performance for each method, where OSR is measured as the AUROC for classifying data as ID or OOD. For our method, FixMatch, and fully supervised, we use the free-energy score \eqref{eq:s-score} to evaluate AUROC, whereas for the others we use the scores proposed in the respective paper.

\subsection{Datasets}

We use several datasets with different characteristics for a complete performance assessment. For ID data, we use CIFAR-10, CIFAR-100 \cite{krizhevsky2009learning} and ImageNet-30 \cite{hendrycks2019using}. The CIFAR sets are of size $32^2$ and comprise 10 and 100 classes, respectively. Both have training sets of 50,000 images and test sets of size 10,000. When CIFAR-10/100 is used as ID, we use SVHN \cite{netzer2011reading}, uniform noise, and the corresponding CIFAR set as OOD. SVHN consists of images showing house numbers divided into 73,257 images in the training set and 26,032 for the test set. The uniform noise dataset has 50,000 training images and 10,000 test images. ImageNet-30 is a 30-class subset of ImageNet, selected such that there is no overlap between the classes. Following \cite{saito2021openmatch}, we use the first 20 classes as ID and the last 10 classes as OOD. Each class has 1,300 training images and 100 test images. The images of ImageNet-30 are first resized so that the shortest side gets a length of 256 while keeping the aspect ratio. Each image is then center-cropped to size $224^2$.

The different open-set scenarios presented above pose distinct challenges. CIFAR-10 and CIFAR-100 originate from the same source set \cite{krizhevsky2009learning} and contain semantically similar classes, \eg, CIFAR-10 contains \emph{dogs} whereas CIFAR-100 has images of \emph{wolves}, making for a challenging OSR problem. %
On the other hand, the similarities could potentially increase the possibility of learning useful features from the OOD set. 
Conversely, we have OOD data constituting pure noise images containing no semantic content or learnable features. However, they could cause unexpected behavior if misclassified and thus generate unwanted training signals. The in-between scenario is the SVHN set containing real images with learnable features but is semantically very different from CIFAR-10 and CIFAR-100. ImageNet-30 showcases how the methods perform on more realistic data similar to that used in practical applications.  %
Example images are shown in \cref{fig:data-examples}.

\subsection{Limitations}

The experiments in this work only consider OSSL problems where the OOD data in training and testing follow the same distributions.  Furthermore, we do not evaluate our method for very low-label regimes (\ie, only a few labeled training samples per class). Lastly, we only use ID sets that are balanced in terms of classes.

\subsection{Implementation details}
\label{seq:implementation}

\begin{table*}[ht!]
    \footnotesize
    \centering
    \setlength{\tabcolsep}{2.0pt}
    \resizebox{1\linewidth}{!}{%
    \input{tables/cifar-all-results}    }
    \vspace*{-0.15cm}
    \caption{Closed-set accuracy (top rows) and AUROC (bottom rows) for SSL and OSSL methods when using CIFAR-10/100 as ID data. Boldface marks the best closed-set accuracies and underline marks the best AUROCs among OSSL methods.}
    \label{tab:cifar-results}
    \vspace*{-0.2cm}
\end{table*}

\begin{table}[]
    \footnotesize
    \centering
    \input{tables/imagenet-30-results}    \vspace*{-0.15cm}
    \caption{Closed-set accuracy and AUROC on ImageNet-30 dataset when using 2,600 labels. Baseline results are taken from \cite{saito2021openmatch}.}
    \label{tab:imagenet-results}
    \vspace*{-0.2cm}
\end{table}

\begin{table}[]
    \footnotesize
    \centering
    \setlength{\tabcolsep}{2.0pt}
    \resizebox{\linewidth}{!}{%
    \input{tables/cifar100-results-additional-data}    }
    \vspace*{-0.15cm}
    \caption{Closed-set accuracy and AUROC with CIFAR-100 as ID using an additional 50 labels/class as validation data. OpenMatch uses the validation data for selecting the best model during training, and SeFOSS uses the data as extra labeled training data.}
    \label{tab:cifar100-results-validation}
    \vspace*{-0.4cm}
\end{table}

The architectures used for the experiments are WRN-28-2 \cite{zagoruyko2016wide} (ID: CIFAR-10), WRN-28-8 \cite{zagoruyko2016wide} (ID: CIFAR-100), and ResNet-18 \cite{he2016deep} (ImageNet-30). In SeFOSS, when CIFAR-10 is ID, we use $w_e=10^{-4},~w_s=5.0,~\eta_0=0.03, \gamma=7/8, B=64, \mu=7, \xi_\text{id} = 0.2, \xi_\text{ood}=1.3, \zeta_\text{ood}=1.9, K_p=5 \cdot 10^4, K=4 \cdot 10^5, w_d = 5 \cdot 10^{-4}$, and SGD momentum 0.9. When CIFAR-100 is ID we use $w_s=15.0, w_d = 10^{-4}$ and $\gamma=5/8$ (following \cite{wallin2022doublematch}), keeping the other hyperparameters the same. For ImageNet-30, we use the same hyperparameters as for CIFAR-10 except that we use $K = 2 \cdot 10^5$ because of the more expensive training steps. We evaluate SeFOSS using an exponential moving average of the model parameters (with momentum 0.999). For T2T and OpenMatch, we use the original authors' implementations and hyperparameters. %
For a fair comparison, we implement MTCF with the FixMatch backbone (with original FixMatch hyperparameters). Our experiments with FixMatch use hyperparameters from the original work. The fully supervised baseline is trained for 50,000 steps, uses batch size 256, and a learning rate from \eqref{eq:lr} with $\gamma=7/8$ and $\eta_0=0.03$.

\subsection{OSSL performance}
\label{seq:results}

{\bf CIFAR-10/100:} The results from using CIFAR-10 and CIFAR-100 as ID data are shown in \cref{tab:cifar-results}. For CIFAR-10, we use labeled sets of sizes 1,000 and 4,000, whereas, for CIFAR-100, the sets have sizes 2,500 and 10,000. The top row for each method shows closed-set accuracy in \%, and the bottom row shows AUROC for OSR. We evaluate each combination of ID and OOD datasets for each method using five different sub-samplings of the complete labeled data. The reported numbers are the mean and standard deviation from these five training sessions. The results from each session are evaluated by taking median performance values from the last five model evaluations during training. 

\cref{tab:cifar-results} shows that SeFOSS is the only method that reaches good performances for both closed-set accuracy and AUROC across all scenarios. MTCF shows overall good AUROC but generally performs worse than SeFOSS on closed-set accuracy. T2T reaches good results on a few scenarios (\eg CIFAR-100 with 10,000 labels with noise as OOD) but does not consistently perform well across scenarios. A drawback of T2T is that it displays high variance in AUROC for many scenarios. OpenMatch shows very good results in terms of closed-set accuracy when CIFAR-10 is ID. OpenMatch however seems unable to handle noise as OOD data since it displays poor and high-variance AUROC for these scenarios. OpenMatch also drops in performance in the CIFAR-100 experiments for both closed-set accuracy and AUROC, where it is outperformed by the fully supervised lower bound for some scenarios. Slightly surprisingly, the closed-set SSL method FixMatch obtains the highest closed-set accuracy in nearly all scenarios. More expected, however, is that FixMatch consistently gives poor AUROC since it can freely assign pseudo-labels to OOD samples, leading to erroneous and overconfident predictions on these data. Existing works \cite{huang2021trash, saito2021openmatch} have reported comparably worse performance for FixMatch, the reason for this is likely the use of hyperparameters that are different from the original work (\eg, fewer training steps).

{\bf ImageNet-30:} For the results on ImageNet-30, we compare with numbers reported by \cite{saito2021openmatch}. The number of labeled data used here is 2,600. To make our results comparable with \cite{saito2021openmatch}, we report the test performance also for SeFOSS at the point of best validation performance given a labeled validation set of 1,000 images. The reported numbers are means and standard deviations over three runs. The results in \cref{tab:imagenet-results} show that SeFOSS reaches better results than MTCF and OpenMatch in terms of both AUROC and closed-set accuracy. Note also that the hyperparameters used for SeFOSS are the same as for CIFAR-10, indicating that SeFOSS scales well to high-resolution data.

{\bf Avoiding collapse using validation data:} To present the fairest possible evaluation of OpenMatch, we note that the poor results displayed in \cref{tab:cifar-results} when CIFAR-100 is ID are many times a result of a training collapse from a much better performance. This collapse can be avoided by using a labeled validation set and selecting the model that yields the best performance on the validation set during training. The official code of \cite{saito2021openmatch} uses 50 images per class for this purpose. The results from evaluating OpenMatch using this procedure are shown in \cref{tab:cifar100-results-validation}, where we see that OpenMatch displays much better results, although it does not solve the poor AUROC for noise OOD. However, for fairness, as SeFOSS does not suffer from training collapse and, thus, has no need for a validation set, it is free to use the additional data during training instead, resulting in a significant boost to closed-set accuracy.  
As SSL methods are meant for situations where labeled data are scarce or expensive, assuming the presence of a sufficiently large labeled validation set during training goes against this philosophy.

\subsection{Influence of OOD data on SSL methods}

From \cref{tab:cifar-results}, we see that FixMatch displays high closed-set accuracies for all datasets. These results contradict prior works where OOD data in SSL are assumed to significantly harm the closed-set performance \cite{guo2020safe, he2022safe}. To investigate this further, we study how a few SSL methods perform when trained using unlabeled data containing different fractions of OOD data. The SSL methods that we evaluate are FixMatch, UDA \cite{xie2019unsupervised}, and MixMatch \cite{berthelot2019mixmatch}. We also include SeFOSS for comparison. The dataset used consists of CIFAR-10 with 4,000 labels as ID data and CIFAR-100 as OOD data. The unlabeled datasets with different fractions of OOD data are created by adding CIFAR-100 data (up to 0.5) or removing CIFAR-10 data (above 0.5). For OSR, the SSL methods are evaluated using the free-energy score \eqref{eq:s-score}. The results are shown in \cref{fig:partial-mixes}.

Most notable from the results is that FixMatch and UDA show no significant drop in closed-set accuracy when the fraction of OOD data is below 0.4. MixMatch loses closed-set accuracy faster, likely due to its use of mixup \cite{zhang2018mixup} augmentations, which intuitively should not handle OOD data well. For AUROC, we see a quick drop in performance for all SSL methods as OOD data gets added to the unlabeled set. It is also here we see a significant difference in the performance of SeFOSS and the traditional SSL methods. Our framework consistently shows high AUROC (around 0.9) when the fraction of OOD data is below 0.7.

\subsection{Ablation}
\label{seq:ablation}

SeFOSS uses three loss functions to learn from unlabeled data, see \cref{fig:method}. To study the importance of these terms, we conduct experiments using 1) only self-supervision on unlabeled data, 2) self-supervision and energy regularization, and 3) self-supervision and pseudo-labeling. Additionally, we evaluate the OSR performance of case 1) using the maximum softmax confidence to confirm that the free-energy score gives better performance.

The results in \cref{tab:ablation} show that using only the self-supervision gives nearly as good results as using the entire framework. Adding pseudo-labeling and energy regularization barely affects the closed-set accuracy but gives better AUROC by a couple of percentage points. This indicates that self-supervision alone, at least under these conditions, is a strong and safe training signal for OSSL. 

Moreover, we study the effect of manually adjusting $\tau_\text{id}$. In \cref{tab:ablation-thresholds} we see an increase in accuracy at the cost of a reduction in AUROC when lowering $\tau_\text{id}$. The increase in accuracy can be explained by the model assigning pseudo-labels to more data. However, when $\tau_\text{id}$ is low, many predicted inliers are likely outliers, causing weaker OSR performance. These experiments are done with $w_e = 0$ to isolate the effect of $\tau_\text{id}$.

\begin{figure}[ht!]
    \centering
    \input{figures/cifar-partial-mixes}%
    \vspace*{-0.2cm}
    \caption{Closed-set accuracy and AUROC using different fractions of OOD data in the unlabeled set. ID data are CIFAR-10 with 4,000 labels. The OOD set is CIFAR-100.}
    \label{fig:partial-mixes}
\end{figure}

\begin{table}[h!]
    \footnotesize
    \centering
    \setlength{\tabcolsep}{2.0pt}
    \input{tables/ablation}
    \vspace*{-0.1cm}
    \caption{Evaluating different modifications of SeFOSS.}
    \label{tab:ablation}
    \vspace*{-0.4cm}
\end{table}

\begin{table}[h!]
    \footnotesize
    \centering
    \setlength{\tabcolsep}{2.0pt}
    \input{tables/ablation-thresholds-cifar100}
    \vspace*{-0.1cm}
    \caption{Evaluating SeFOSS when manually adjusting $\tau_\text{id}$.}
    \label{tab:ablation-thresholds}
    \vspace*{-0.5cm}
\end{table}

\section{Conclusion}

This paper shows that self-supervision on all unlabeled data combined with OSR predictions using the free-energy score can be successfully applied in an OSSL context. Our proposed framework reaches overall strong and consistent results on a wide range of OSSL problems when considering both closed-set accuracy and OSR. In particular, it displays SOTA performance on the more challenging tests when CIFAR-100 is ID and the more realistic scenarios using ImageNet-30. However, we should note that the performance is marginally below SOTA when CIFAR-10 is ID. Moreover, if focusing solely on closed-set accuracy, we show that SSL methods can perform equal to or better than designated OSSL methods even when unlabeled data contain OOD data, indicating that the challenge with OSSL lies in ensuring OSR performance during employment.

\section*{Acknowledgment}

This work was supported by Saab AB, the Swedish Foundation for Strategic Research, and Wallenberg AI, Autonomous Systems and Software Program (WASP) funded by the Knut and Alice Wallenberg Foundation.
The experiments were enabled by resources provided by the Swedish National Infrastructure for Computing (SNIC) at Chalmers Centre for Computational Science and Engineering (C3SE), and National Supercomputer Centre (NSC) at Linköping University.

{\small
\bibliographystyle{ieee_fullname}
\bibliography{mybib}
}

\end{document}


\title{Supplementary Material:\\Improving Open-Set Semi-Supervised Learning with Self-Supervision}

\author{Erik Wallin\textsuperscript{1,2}, Lennart Svensson\textsuperscript{2}, Fredrik Kahl\textsuperscript{2}, Lars Hammarstrand\textsuperscript{2} \\
\textsuperscript{1}Saab AB, \textsuperscript{2}Chalmers University of Technology \\
\{walline,lennart.svensson,fredrik.kahl,lars.hammarstrand\}@chalmers.se}

\maketitle

\section{Choices of hyperparameters}

While SeFOSS may appear to have an extensive set of hyperparameters, most values are gathered from existing works and have been used successfully without further tuning. For example, the architectures for the CIFAR experiments follow FixMatch \cite{sohn2020fixmatch} and the architecture for ImageNet-30 follows OpenMatch \cite{saito2021openmatch}. Parameters such as the labeled and unlabeled batch sizes, weight regularization, SGD momentum, EMA momentum, learning rate, and the learning rate schedule all follow FixMatch and have not been further tuned for SeFOSS. However, the initial constant learning rate for the pre-training phase is a new addition for SeFOSS.

The decay rates for the learning rates for CIFAR-10 and CIFAR-100 follow DoubleMatch \cite{wallin2022doublematch} without further tuning for SeFOSS. The temperature scaling of the energy score is set to $\beta=1$ as suggested by the original work \cite{li2020energy}. The weight for the self-supervised feature loss, $w_s$, was largely determined based on results from DoubleMatch, which found that a smaller $w_s$ works better for CIFAR-10 and a larger $w_s$ works better for CIFAR-100. The number of training steps is empirically determined and is not a very sensitive hyperparameter; no large performance gains were found by increasing the number of training steps further. The weight for the energy regularization, $w_e$, is new for SeFOSS and was determined by grid searches using CIFAR-10 as ID and CIFAR-100 as OOD. The hyperparameters determining the thresholds for pseudo inliers and outliers, and the margin for the hinge loss  ($\xi_\text{id}$, $\xi_\text{ood}$ and $\zeta_\text{ood}$), were initially given rough estimates by observing the empirical distributions of energies for ID and OOD test data in relation the the distribution of energies for labeled ID training data. These hyperparameters were then more carefully tuned by grid search using CIFAR-10 as ID and CIFAR-100 as OOD.

\section{Motivating the free-energy score}

The original work that proposes the free-energy score for OOD detection \cite{li2020energy} finds that the free-energy score generally outperforms the baseline of using the maximum softmax confidence, $\max_y p(y|x)$. To further motivate the use of the free-energy score in SeFOSS, we conduct experiments where we evaluate AUROC for the confidence score and for the free-energy score at the end of the pre-training phase of SeFOSS. In the pre-training phase, the only losses used are the supervised cross-entropy, $l_l$, the self-supervised feature consistency, $l_s$, and the weight regularization, $l_w$. We perform these evaluations at the end of the pre-training phase, because until then, the energy regularization that amplifies the performance of the energy-score is not applied. The results are shown in \cref{tab:conf-vs-energy}. We see that the energy score performs better or equal to the confidence score in all scenarios, giving strong support to the design choice of using the free-energy score in SeFOSS.

\begin{table*}[ht!]
    \footnotesize
    \centering
    \setlength{\tabcolsep}{2.0pt}
    \resizebox{1\linewidth}{!}{%
    \input{tables/conf-vs-energy}%
    }
    \vspace*{-0.15cm}
    \caption{AUROC at the end of the pretraining phase for SeFOSS using either the softmax confidence or the free-energy score.}
    \label{tab:conf-vs-energy}
    \vspace*{-0.2cm}
\end{table*}

\section{OSR for previously unseen OOD}

The results in the main paper evaluate open-set recognition at testing for the OOD classes that appear in the unlabeled set. While we argue that this scenario is closest to real-world applications, it can also be of interest to evaluate the performance for open-set recognition on classes that are completely unseen during training. For example, the model may be trained to classify CIFAR-10, encounters OOD data from CIFAR-100 in the unlabeled training data, but is then evaluated for open-set recognition with CIFAR-10 as ID and SVHN as OOD. Results from these experiments are shown in \cref{tab:cifar10-4000-cifar100,tab:cifar10-4000-noise,tab:cifar10-4000-svhn,tab:cifar100-10000-cifar10,tab:cifar100-10000-svhn,tab:cifar100-10000-noise}. We see that SeFOSS generally achieves high AUROCs also for unseen OOD data. The scenarios where we observe poor performance are when we evaluate using CIFAR-10 as unseen OOD when CIFAR-100 is ID. This seems reasonable because many of the classes in CIFAR-10 are very similar to classes that appear in CIFAR-100, and the model has had no chance to learn to distinguish between these in training.

\begin{table}[]
    \centering
    \begin{tabular}{cc c}
    \toprule
      Test OOD & SVHN & Noise \\ \midrule
      AUROC & 0.99 & 0.98 \\ \bottomrule
    \end{tabular}
    \caption{ID: CIFAR-10 4,000 labels, Train OOD: CIFAR-100}
    \label{tab:cifar10-4000-cifar100}
\end{table}

\begin{table}[]
    \centering
    \begin{tabular}{cc c}
    \toprule
      Test OOD & CIFAR-100 & Noise \\ \midrule
      AUROC & 0.90 & 0.99 \\ \bottomrule
    \end{tabular}
    \caption{ID: CIFAR-10 4,000 labels, Train OOD: SVHN}
    \label{tab:cifar10-4000-svhn}
\end{table}

\begin{table}[]
    \centering
    \begin{tabular}{cc c}
    \toprule
      Test OOD & CIFAR-100 & SVHN \\ \midrule
      AUROC & 0.91 & 0.95 \\ \bottomrule
    \end{tabular}
    \caption{ID: CIFAR-10 4,000 labels, Train OOD: Noise}
    \label{tab:cifar10-4000-noise}
\end{table}

\begin{table}[]
    \centering
    \begin{tabular}{cc c}
    \toprule
      Test OOD & SVHN & Noise \\ \midrule
      AUROC & 0.91 & 0.94 \\ \bottomrule
    \end{tabular}
    \caption{ID: CIFAR-100 10,000 labels, Train OOD: CIFAR-10}
    \label{tab:cifar100-10000-cifar10}
\end{table}

\begin{table}[]
    \centering
    \begin{tabular}{cc c}
    \toprule
      Test OOD & CIFAR-10 & Noise \\ \midrule
      AUROC & 0.76 & 0.85 \\ \bottomrule
    \end{tabular}
    \caption{ID: CIFAR-100 10,000 labels, Train OOD: SVHN}
    \label{tab:cifar100-10000-svhn}
\end{table}

\begin{table}[]
    \centering
    \begin{tabular}{cc c}
    \toprule
      Test OOD & CIFAR-10 & SVHN \\ \midrule
      AUROC & 0.77 & 0.92 \\ \bottomrule
    \end{tabular}
    \caption{ID: CIFAR-100 10,000 labels, Train OOD: Noise}
    \label{tab:cifar100-10000-noise}
\end{table}

\newpage

{\small
\bibliographystyle{ieee_fullname}
\bibliography{mybib}
}

%% file: figures/introfig-ossl.tex
\begin{tikzpicture}[,
rectnode/.style={rectangle, draw=black, align=center}
]

\footnotesize

\newcommand{\myfigsize}{0.08\textwidth}

\pgfdeclarelayer{bg1}    %
\pgfdeclarelayer{bg2}
\pgfsetlayers{bg1,bg2,main}  %

\node (lb1) [inner sep=0pt] {\includegraphics[width=\myfigsize]{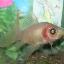}};
\node (lb2) [inner sep=0pt, below=0.1cm of lb1] {\includegraphics[width=\myfigsize]{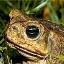}};

\node (ssl-ul1) [inner sep=0pt, right=0.4cm of lb1] {\includegraphics[width=\myfigsize]{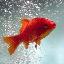}};
\node (ssl-ul2) [inner sep=0pt, below=0.1cm of ssl-ul1] {\includegraphics[width=\myfigsize]{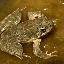}};
\node (ossl-ul1) [inner sep=0pt, right=0.6cm of ssl-ul1] {\includegraphics[width=\myfigsize]{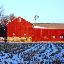}};
\node (ossl-ul2) [inner sep=0pt, below=0.1cm of ossl-ul1] {\includegraphics[width=\myfigsize]{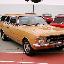}};

\node (labeledtext) [align=center, anchor=south] at (lb1.north) {Labeled};

\node (ssl-ultext) [align=center, anchor=south] at (ssl-ul1.north) {Unlabeled ID};

\node (ossl-ultext) [align=center, anchor=south] at (ossl-ul1.north) {Unlabeled OOD};

\node (top-left-ossl) [inner sep=0pt, outer sep=0pt] at ($(lb1.north west)+(-0.6cm,+0.9cm)$) {};
\node (bottom-right-ossl) [inner sep=0pt, outer sep=0pt] at ($(ossl-ul2.south east)+(0.6cm,-0.4cm)$) {};

\node (top-left-ssl) [inner sep=0pt, outer sep=0pt] at ($(lb1.north west)+(-0.2cm,+0.4cm)$) {};
\node (bottom-right-ssl) [inner sep=0pt, outer sep=0pt] at ($(ssl-ul2.south east)+(0.2cm,-0.2cm)$) {};

\begin{pgfonlayer}{bg2}
    \node (ssl-rect) [draw=black, inner sep=0pt, outer sep=0pt, shade, shading=axis, left color=white, right color=lightgray, shading angle=90, fit={(top-left-ssl)(bottom-right-ssl)}] {};
\end{pgfonlayer}

\node (ssl-text) [align=center, above=0cm of ssl-rect.north] {Closed-set semi-supervised learning};

\begin{pgfonlayer}{bg1}
    \node (ossl-rect) [draw=black, inner sep=0pt, outer sep=0pt, shade, shading=axis, left color=white, right color=lightgray, shading angle=90, fit={(top-left-ossl)(bottom-right-ossl)}] {};
\end{pgfonlayer}

\node (ossl-text) [align=center, above=0cm of ossl-rect.north] {Open-set semi-supervised learning};

\end{tikzpicture}

%% file: figures/method-graph.tex
\begin{tikzpicture}[,
rectnode/.style={rectangle, draw=black, align=center}
]

\newcommand{\myfigsize}{0.052\textwidth}

\footnotesize

\node (uimg1) [inner sep=0pt] at (-2, 2) {\includegraphics[width= \myfigsize]{images/n01443537_0.JPEG}};

\node (uimg2) [inner sep=0pt, below=0.1 cm of uimg1] {\includegraphics[width= \myfigsize]{images/n01443537_10.JPEG}};

\node (uimg3) [inner sep=0pt, below=0.1 cm of uimg2] {\includegraphics[width= \myfigsize]{images/n01644900_0.JPEG}};

\node (uimg4) [inner sep=0pt, right=0.1 cm of uimg1] {\includegraphics[width= \myfigsize]{images/n02793495_7.JPEG}};

\node (uimg5) [inner sep=0pt, below=0.1 cm of uimg4] {\includegraphics[width= \myfigsize]{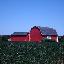}};

\node (uimg6) [inner sep=0pt, below=0.1 cm of uimg5] {\includegraphics[width= \myfigsize]{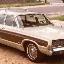}};

\node (model) [trapezium, draw=black, shape border rotate=270, shade, shading=axis, left color=white, right color=teal, shading angle=90, right=0.5 cm of uimg5] {$s \circ g \circ f$};

\node (uimgtext) [rotate=90] at ($(uimg2.west) + (-0.2cm,0)$) {Unlabeled data};

\node (oimg1) [inner sep=0pt, right=1.5cm of model] {\includegraphics[width= \myfigsize]{images/n01443537_10.JPEG}};

\node (oimg2) [inner sep=0pt, right=0.1cm of oimg1] {\includegraphics[width= \myfigsize]{images/n02793495_7.JPEG}};

\node (pol1) [inner sep=0pt, above=0.35cm of oimg1] {\includegraphics[width= \myfigsize]{images/n02793495_0.JPEG}};

\node (pol2) [inner sep=0pt, right=0.1cm of pol1] {\includegraphics[width= \myfigsize]{images/n02814533_0.JPEG}};

\node (pil1) [inner sep=0pt, below=0.35cm of oimg1] {\includegraphics[width= \myfigsize]{images/n01644900_0.JPEG}};

\node (pil2) [inner sep=0pt, right=0.1cm of pil1] {\includegraphics[width= \myfigsize]{images/n01443537_0.JPEG}};

\node (oimgtext) [anchor=south, align=center] at ($(oimg1.north)!0.5!(oimg2.north)$) {Uncertain data};
\node (poltext) [anchor=south, align=center] at ($(pol1.north)!0.5!(pol2.north)$) {Pseudo outliers};
\node (piltext) [anchor=south, align=center] at ($(pil1.north)!0.5!(pil2.north)$) {Pseudo inliers};

\node (ssl) [align=center, right=of oimg2] {$l_s$};
\node (ood) [align=center, right=0.5 cm of ssl] {$l_e$};
\node (pl) [align=center, right=0.5 cm of ood] {$l_{p}$};

\draw[->] (uimg5) -- (model);

\draw[->] (model) |- (pol1) node [near end, above] {$>\tau_\text{ood}$};
\draw[->] (model) -- (oimg1) node [midway, above] {$<\tau_\text{ood}$} node [midway, below] {$>\tau_\text{id}$};
\draw[->] (model) |- (pil1) node [near end, below] {$<\tau_\text{id}$};

\draw[->] (oimg2) -- (oimg2-|ssl.west);
\draw[->] (pol2) -| (ssl);
\draw[->] (pil2) -| (ssl);

\draw[->] (pol2) -| (ood);

\draw[->] (pil2) -| (pl);

\end{tikzpicture}

%% file: algorithms/trainingloop.tex
\algnewcommand{\LineComment}[1]{\State \(\triangleright\) #1}
\begin{algorithmic}[1]
    \Require Trainable models $f$, $g$, and $h$, labeled training data $\{(\mathbf{x}_1, \mathbf{y}_1),\dots,(\mathbf{x}_M, \mathbf{y}_M)\}$, unlabeled training data $\{\tilde{\mathbf{x}}_1, \dots, \tilde{\mathbf{x}}_{\tilde{M}} \}$,  scaling parameters $w_p$, $w_s$, $w_e$, and $w_w$, learning rate scheme $\eta(k)$
    
    \Statex \vspace{-.7em}
    \LineComment{Pretraining loop}
    \For{$i = 1,\dots, K_\text{p}$}
        \State Compute loss according to \cref{alg:energymatch} setting $w_e = w_p = 0$
        \State Update parameters of $f$, $g$, and $h$ using SGD%
    \EndFor
    \Statex \vspace{-.7em}
    \LineComment{Compute confidence thresholds by looping through all labeled data}
    \For{$i = 1, \dots, M$}
        \State $S_i = s( g ( f (\mathbf{x}_i)))$
    \EndFor
    \State Compute thresholds and margin $\tau_\text{id}$, $\tau_\text{ood}$, and $m_\text{ood}$ from \cref{eq:thresholds}
    \Statex \vspace{-.7em}
    \LineComment{Training loop}
    \For{$i = K_\text{p}+1, \dots, K$} 
        \State Compute loss using \cref{alg:energymatch} using $w_p=1$ and a tuned $w_e$%
        \State Update parameters of $f$, $g$, and $h$ using SGD%
    \EndFor \vspace{1em}
    \Statex \vspace{-.7em}
    \Return trained classification model $g \circ f$
\end{algorithmic}

%% file: algorithms/energymatch.tex
\algnewcommand{\LineComment}[1]{\State \(\triangleright\) #1}
\begin{algorithmic}[1]
    \Require Strong augmentation $\beta$, weak augmentation $\alpha$, labeled batch $\{(\mathbf{x}_1, \mathbf{y}_1),\dots,(\mathbf{x}_B,\mathbf{y}_B)\}$, unlabeled batch $\{\tilde{\mathbf{x}}_1,\dots,\tilde{\mathbf{x}}_{\mu B}\}$, scaling parameters $w_p$, $w_s$, $w_e$, and $w_w$, thresholds $\tau_\text{id}$, $\tau_\text{ood}$, and $m_\text{ood}$, backbone model $f$, prediction layer $g$, projection layer $h$
    
    \Statex \vspace{-.7em}
    \LineComment{Cross-entropy loss for (weakly augmented) labeled data}
    \For{$i = 1,\cdots, B$} 
        \State $\mathbf{o}_i = g ( f (\alpha(\mathbf{x}_i)))$
    \EndFor
    \State $l_l = \frac{1}{B} \sum_{i=1}^B H(\mathbf{y}_i,\softmax(\mathbf{o}_i))$
    \Statex \vspace{-.7em}
    \LineComment Predictions on unlabeled data
    \For{$i = 1, \cdots, \mu B$} 
        \State $\mathbf{z}_i = f(\alpha(\tilde{\mathbf{x}}_i))$ \Comment{Weak augmentation}
        \State $\mathbf{v}_i = f(\beta(\tilde{\mathbf{x}}_i))$ \Comment{Strong augmentation}
        \State $\mathbf{q}_i = g(\mathbf{v}_i)$
        \State $\mathbf{w}_i = g(\mathbf{z}_i)$
    \EndFor
    
    \State Compute $l_s$, $l_p$, $l_e$, and $l_w$ according to \eqref{eq:ls}, \eqref{eq:lp}, \eqref{eq:le}, and \eqref{eq:weight-decay} \vspace{1em}
    \Statex \vspace{-.7em}
    \Return $l_l + w_p l_p + w_s l_s + w_e l_e + w_w l_w$
\end{algorithmic}

%% file: figures/data-examples.tex
\begin{tikzpicture}[,
rectnode/.style={rectangle, draw=black, align=center}
]

\newcommand{\mywidthsep}{0.3 cm}
\newcommand{\myheightsep}{0.5 cm}
\newcommand{\myfigsize}{0.08\textwidth}
\newcommand{\mysmallsep}{0.08 cm}

\footnotesize

\node (cifar10-1) [inner sep=0pt] at (0,0) {\includegraphics[width=\myfigsize]{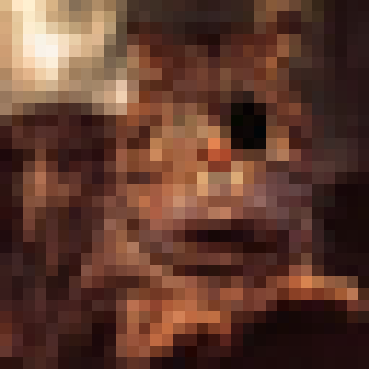}};
\node (cifar10-2) [inner sep=0pt, right=\mysmallsep of cifar10-1] {\includegraphics[width=\myfigsize]{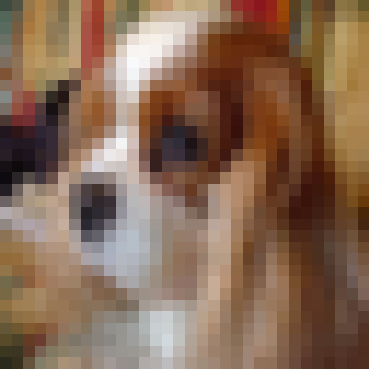}};

\node (cifar10-text) [align=center, above=0cm of $(cifar10-1.north)!0.5!(cifar10-2.north)$] {CIFAR-10};

\node (cifar100-1) [inner sep=0pt, right=\mywidthsep of cifar10-2] {\includegraphics[width=\myfigsize]{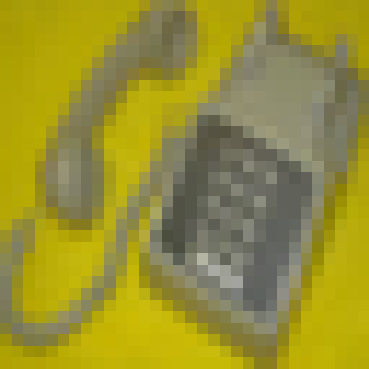}};
\node (cifar100-2) [inner sep=0pt, right=\mysmallsep of cifar100-1] {\includegraphics[width=\myfigsize]{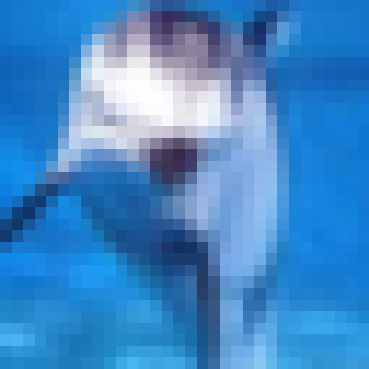}};

\node (cifar100-text) [align=center, above=0cm of $(cifar100-1.north)!0.5!(cifar100-2.north)$] {CIFAR-100};

\node (svhn-1) [inner sep=0pt, right=\mywidthsep of cifar100-2] {\includegraphics[width=\myfigsize]{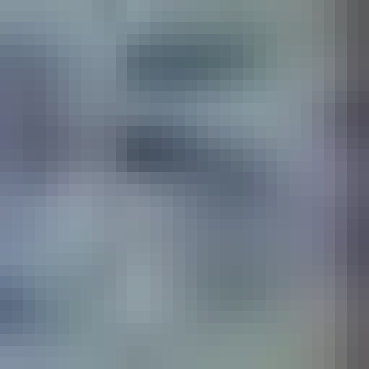}};
\node (svhn-2) [inner sep=0pt, right=\mysmallsep of svhn-1] {\includegraphics[width=\myfigsize]{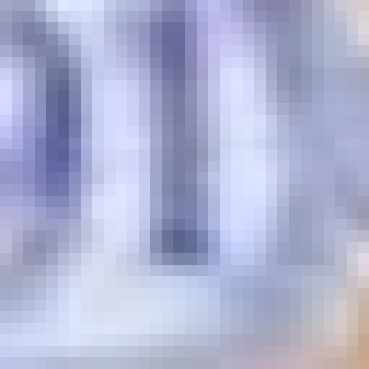}};

\node (imagenet-1) [inner sep=0pt, right=\mywidthsep of svhn-2] {\includegraphics[width=\myfigsize]{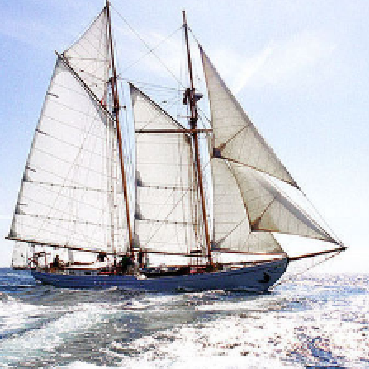}};
\node (imagenet-2) [inner sep=0pt, right=\mysmallsep of imagenet-1] {\includegraphics[width=\myfigsize]{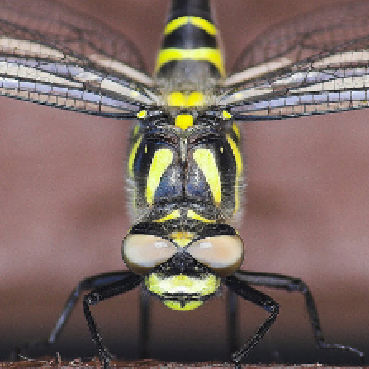}};

\node (imagenet-text) [align=center, above=0cm of $(imagenet-1.north)!0.5!(imagenet-2.north)$] {ImageNet-30};

\node (svhn-text) [align=center, above=0cm of $(svhn-1.north)!0.5!(svhn-2.north)$] {SVHN};

\node (noise-1) [inner sep=0pt, right=\mywidthsep of imagenet-2] {\includegraphics[width=\myfigsize]{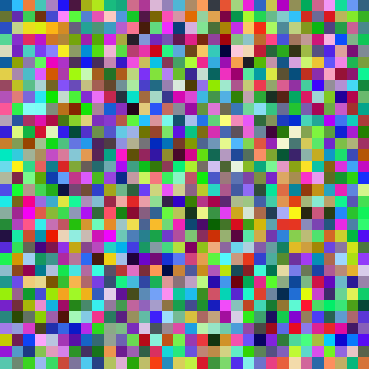}};
\node (noise-2) [inner sep=0pt, right=\mysmallsep of noise-1] {\includegraphics[width=\myfigsize]{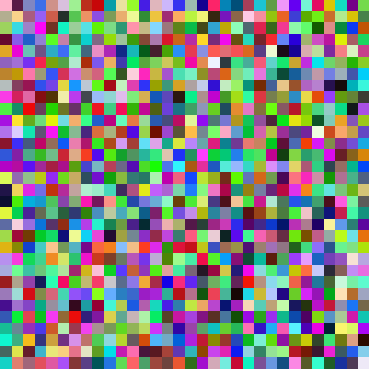}};

\node (noise-text) [align=center, above=0cm of $(noise-1.north)!0.5!(noise-2.north)$] {Uniform noise};

\end{tikzpicture}

%% file: tables/cifar-all-results.tex
\begin{tabular}{@{}r c c c c c c c c c c c c @{}}
\toprule

& \multicolumn{3}{c}{CIFAR-10: 1,000 labels} & \multicolumn{3}{c}{CIFAR-10: 4,000 labels} 
& \multicolumn{3}{c}{CIFAR-100: 2,500 labels} & \multicolumn{3}{c}{CIFAR-100: 10,000 labels}\\
& CIFAR-100 & SVHN & Noise & CIFAR-100 & SVHN & Noise 
& CIFAR-10  & SVHN & Noise & CIFAR-10  & SVHN & Noise \\
\cmidrule(lr){2-4} \cmidrule(lr){5-7} \cmidrule(lr){8-10} \cmidrule(lr){11-13}

\multirow{2}{*}{Fully supervised} 
& 54.51\spm{1.82} & 54.03\spm{2.05} & 55.39\spm{2.67} & 75.57\spm{2.88} & 76.70\spm{2.27} & 77.62\spm{1.79} 
& 34.62\spm{1.43} & 33.19\spm{1.80} & 34.03\spm{1.50} & 59.12\spm{0.91} & 60.32\spm{0.54} & 59.40\spm{2.11}\\
& 0.62\spm{0.01} & 0.61\spm{0.04} & 0.56\spm{0.22} & 0.74\spm{0.02} & 0.80\spm{0.03}  & 0.78\spm{0.15} 
& 0.61\spm{0.01} & 0.57\spm{0.05} & 0.34\spm{0.19} & 0.71\spm{0.01} & 0.76\spm{0.08} & 0.77\spm{0.17}\\ 
\midrule 

\multirow{2}{*}{FixMatch \cite{sohn2020fixmatch}} 
& 92.70\spm{0.14} & 94.80\spm{0.19} & 95.02\spm{0.10} & 94.07\spm{0.15} & 94.93\spm{0.22} & 95.38\spm{0.07} 
& 71.95\spm{0.49} & 69.39\spm{0.14} & 70.89\spm{0.42} & 77.72\spm{0.32} & 75.89\spm{0.39}  & 77.04\spm{0.24} \\
& 0.66\spm{0.00} & 0.67\spm{0.03} & 0.71\spm{0.05} & 0.69\spm{0.01} & 0.67\spm{0.02}  & 0.73\spm{0.01} 
& 0.46\spm{0.01} & 0.49\spm{0.03} & 0.67\spm{0.17} & 0.51\spm{0.01} & 0.48\spm{0.03} & 0.65\spm{0.04}\\ 
\midrule %

\multirow{2}{*}{MTCF \cite{yu2020multi}} 
& 82.96\spm{1.08} & 90.49\spm{0.79} & 89.32\spm{0.65} & 89.87\spm{0.21} & 92.72\spm{0.14} & 92.01\spm{0.32} 
& 40.46\spm{1.49} & 53.55\spm{1.24}  & 46.56\spm{0.66} & 62.88\spm{0.92} & 66.10\spm{0.63} & 63.80\spm{0.75} \\
& 0.81\spm{0.00} & \underline{1.00}\spm{0.00} & \underline{1.00}\spm{0.00} & 0.84\spm{0.00} & \underline{1.00}\spm{0.00}  & \underline{1.00}\spm{0.00} 
& \underline{0.82}\spm{0.01} & \underline{1.00}\spm{0.00} & \underline{1.00}\spm{0.00} & 0.80\spm{0.01} & \underline{1.00}\spm{0.00} & \underline{1.00}\spm{0.00} \\ 
\cdashlinelr{2-13}

\multirow{2}{*}{T2T \cite{huang2021trash}} & 
86.99\spm{1.09} & 91.83\spm{1.20} & 91.13\spm{1.12} & 86.11\spm{1.91} & 92.16\spm{1.00} & 92.91\spm{0.57}
& 38.30\spm{9.72} & 58.44\spm{18.14} & 51.33\spm{9.59} & 62.02\spm{3.73}  & 70.93\spm{4.38} & 73.01\spm{0.37}\\
& 0.57\spm{0.02} & 0.96\spm{0.07} & 0.72\spm{0.26} & 0.57\spm{0.04}  & 0.80\spm{0.24} & 0.90\spm{0.19}
& 0.63\spm{0.08} & 0.80\spm{0.40} & \underline{1.00}\spm{0.00} & 0.59\spm{0.08}  & 0.66\spm{0.42} & \underline{1.00}\spm{0.00} \\ 
\cdashlinelr{2-13}

\multirow{2}{*}{OpenMatch \cite{saito2021openmatch}} %
& \textbf{92.20}\spm{0.15} & \textbf{94.12}\spm{0.34} &\textbf{94.07}\spm{0.08} & \textbf{94.82}\spm{0.21}  & \textbf{94.73}\spm{0.10}  & \textbf{94.76}\spm{0.15} &
20.84\spm{8.65} & 18.66\spm{2.59} & 16.10\spm{5.70} & 40.95\spm{20.44} & 32.69\spm{9.68}  & 21.19\spm{8.55}\\
& \underline{0.93}\spm{0.00}  & 0.98\spm{0.03} &  0.68\spm{0.40} & \underline{0.96}\spm{0.00} &  \underline{1.00}\spm{0.00}  & 0.58\spm{0.24} 
& 0.66\spm{0.05}  & 0.69\spm{0.10} &  0.85\spm{0.18} & 0.77\spm{0.11} &  0.68\spm{0.19}  & 0.50\spm{0.32}\\ 
\cdashlinelr{2-13}

\multirow{2}{*}{SeFOSS (\emph{ours})} 
& 91.49\spm{0.16} & 91.16\spm{0.27} & 92.78\spm{1.00} & 93.73\spm{0.27} & 92.60\spm{0.40} & 94.14\spm{0.09} 
& {\bf 68.48}\spm{0.26} & {\bf 62.99}\spm{0.39} & {\bf 64.54}\spm{1.00} & {\bf 77.63}\spm{0.21} & \textbf{73.60}\spm{0.20}  & \textbf{75.25}\spm{0.34}\\
& 0.90\spm{0.01} & 0.99\spm{0.01} & \underline{1.00}\spm{0.00} & 0.92\spm{0.00} & \underline{1.00}\spm{0.00} & \underline{1.00}\spm{0.00} 
& 0.79\spm{0.01} & \underline{1.00}\spm{0.00} & \underline{1.00}\spm{0.00} & \underline{0.83}\spm{0.00} & \underline{1.00}\spm{0.00} & \underline{1.00}\spm{0.00}\\

\bottomrule
\end{tabular}

%% file: tables/imagenet-30-results.tex
\begin{tabular}{@{}r c c c @{}}
\toprule
& MTCF \cite{yu2020multi} & OpenMatch \cite{saito2021openmatch} & SeFOSS {\em (ours)} \\
\midrule
Accuracy & 86.4\spm{0.7} & 89.6\spm{1.0}  & \textbf{92.53}\spm{0.10}\\
AUROC & 0.94\spm{0.00} & 0.96\spm{0.00} & \underline{0.97}\spm{0.00} \\
\bottomrule
\end{tabular}

%% file: tables/cifar100-results-additional-data.tex
\begin{tabular}{@{}r c c c c c c @{}}
\toprule

& \multicolumn{3}{c}{CIFAR-100: 2,500 labels} & \multicolumn{3}{c}{CIFAR-100: 10,000 labels}\\
& CIFAR-10  & SVHN & Noise & CIFAR-10  & SVHN & Noise \\
\cmidrule(lr){2-4} \cmidrule(lr){5-7}
\multirow{2}{*}{OpenMatch} %
& 63.33\spm{0.86} & 63.41\spm{1.32} & 58.97\spm{0.52}&  75.89\spm{0.23} & 75.56\spm{0.17} & 75.08\spm{0.28}
\\
& \underline{0.86}\spm{0.01} & \underline{1.00}\spm{0.00} & 0.42\spm{0.47} & \underline{0.92}\spm{0.01} & \underline{1.00}\spm{0.00} & 0.24\spm{0.20}\\ 
\cdashlinelr{2-7}

\multirow{2}{*}{SeFOSS} 
& \textbf{76.16}\spm{0.39} & \textbf{71.78}\spm{0.27} & \textbf{73.26}\spm{0.69} & \textbf{79.37}\spm{0.21} & \textbf{ 76.11}\spm{0.28} & \textbf{ 77.53}\spm{0.17} \\
&  0.83\spm{0.01}  & \underline{1.00}\spm{0.00}  & \underline{1.00}\spm{0.00}  & 0.83\spm{0.01}  & \underline{1.00}\spm{0.00}  & \underline{1.00}\spm{0.00} \\ 

\bottomrule
\end{tabular}

%% file: figures/cifar-partial-mixes.tex
\begin{tikzpicture}

    \newcommand{\myheight}{0.19\textwidth}

    \begin{groupplot}[
        group style={
            group size=1 by 2,
            y descriptions at=edge left,
            x descriptions at=edge bottom,
            horizontal sep=23pt,
            vertical sep=5pt,
        },
    ]
    \nextgroupplot[
        title={},
        width=0.4\textwidth,
        height=\myheight,
        ylabel={Accuracy {[\%]}},
        ylabel style={alias=auy},
        xlabel={},
        xmin=0, xmax=1,
        ymin=55, ymax=100,
        legend pos=south west,
        legend columns=2,
        legend style={font=\footnotesize},
        ymajorgrids=true,
        label style={font=\footnotesize},
        tick label style={font=\footnotesize},
        grid style=dashed,
    ]

        \addplot[color=blue,]
        table[col sep=comma, x=percentage, y expr=\thisrow{acc}*100] {figure-data/energymatch.csv};
        
        \addplot[color=red,]
        table[col sep=comma, x=percentage, y expr=\thisrow{acc}*100] {figure-data/fixmatch.csv};
        
        \addplot[color=brown,]
        table[col sep=comma, x=percentage, y expr=\thisrow{acc}*100] {figure-data/mixmatch.csv};
        
        \addplot[color=teal,]
        table[col sep=comma, x=percentage, y expr=\thisrow{acc}*100] {figure-data/uda.csv}; 
        
        \legend{SeFOSS, FixMatch, MixMatch, UDA}

    \nextgroupplot[
        title={},
        width=0.4\textwidth,
        height=\myheight,
        xlabel={Fraction of OOD data in unlabeled training set},
        ylabel={OSR [AUROC]},
        xmin=0, xmax=1,
        ymin=0.45, ymax=1,
        ymajorgrids=true,
        label style={font=\footnotesize},
        tick label style={font=\footnotesize},
        grid style=dashed
    ]
        
        \addplot[color=blue,]
        table[col sep=comma, x=percentage, y=auroc] {figure-data/energymatch.csv};
        
        \addplot[color=red,]
        table[col sep=comma, x=percentage, y=auroc] {figure-data/fixmatch.csv};
        
        \addplot[color=brown,]
        table[col sep=comma, x=percentage, y=auroc] {figure-data/mixmatch.csv};
        
        \addplot[color=teal,]
        table[col sep=comma, x=percentage, y=auroc] {figure-data/uda.csv};   
        
    \end{groupplot}
    
\end{tikzpicture}

%% file: tables/ablation.tex
\begin{tabular}{c c c c c c}
\toprule
\multicolumn{6}{c}{CIFAR-10: 4,000 labels - OOD: CIFAR-100} \\ \midrule
$l_s$ & $l_e$ & $l_p$ & OSR-score & Accuracy & AUROC \\ \midrule
\checkmark & & & conf & 93.87\spm{0.17} & 0.89\spm{0.00} \\
\checkmark & & & energy & 93.87\spm{0.17} & 0.90\spm{0.01} \\
\checkmark & & \checkmark & energy & 93.74\spm{0.17} & 0.90\spm{0.01} \\
\checkmark & \checkmark & & energy & 93.85\spm{0.12} & 0.91\spm{0.01} \\
\checkmark & \checkmark & \checkmark & energy & 93.73\spm{0.27} & 0.92\spm{0.00} \\ \bottomrule
\end{tabular}

%% file: tables/ablation-thresholds-cifar100.tex
\begin{tabular}{ccccccc}
\toprule
\multicolumn{7}{c}{CIFAR-100: 2,500 labels, OOD: CIFAR-10} \\ \midrule
$\tau_\text{id}$ & -30 & -25 & -20 & -15 & -10 & -5 \\ \midrule
Accuracy & 68.93 & 68.44 & 68.51 & 68.35 & 70.21 & 69.25 \\
AUROC & 0.78 & 0.77 & 0.79 & 0.78 & 0.68 & 0.57 \\
\bottomrule
\end{tabular}

%% file: tables/conf-vs-energy.tex
\begin{tabular}{@{}r c c c c c c c c c c c c @{}}
\toprule

& \multicolumn{3}{c}{CIFAR-10: 1,000 labels} & \multicolumn{3}{c}{CIFAR-10: 4,000 labels} 
& \multicolumn{3}{c}{CIFAR-100: 2,500 labels} & \multicolumn{3}{c}{CIFAR-100: 10,000 labels}\\
& CIFAR-100 & SVHN & Noise & CIFAR-100 & SVHN & Noise 
& CIFAR-10  & SVHN & Noise & CIFAR-10  & SVHN & Noise \\
\cmidrule(lr){2-4} \cmidrule(lr){5-7} \cmidrule(lr){8-10} \cmidrule(lr){11-13}

Confidence & 0.83 & 0.99 & 1.0 & 0.84 & 0.98 & 1.0 & 0.64 & 0.98 & 1.00 & 0.72 & 0.99 & 1.00 \\
Energy & 0.84 & 1.00 & 1.0 & 0.86 & 1.00 & 1.0 & 0.66 & 1.00 & 1.00 & 0.74 & 1.00 & 1.00 \\

\bottomrule
\end{tabular}